\pdfoutput=1
\documentclass[11pt,a4]{article}
\usepackage[hyperref]{acl2020}
\usepackage{times}
\usepackage{latexsym}
\usepackage{rotating}
\usepackage{pifont}
\newcommand{\cmark}{\ding{51}}

\usepackage{microtype}
\usepackage{multirow}
\usepackage{amssymb}
\usepackage{subcaption}
\usepackage[hang,flushmargin]{footmisc}  

\aclfinalcopy 
\def\aclpaperid{79} 

\newcommand{\LN}{\linebreak\noindent}
\newcommand{\MC}[2]{\multicolumn{1}{#1}{\textbf{#2}}}

\title{Towards Unified Dialogue System Evaluation: \\ A Comprehensive Analysis of Current Evaluation Protocols}

\author{Sarah E. Finch \\
  Department of Computer Science \\
  Emory University \\
  Atlanta, GA, USA \\
  \texttt{sfillwo@emory.edu} \\\And
  Jinho D. Choi \\
  Department of Computer Science \\
  Emory University \\
  Atlanta, GA, USA \\
  \texttt{jinho.choi@emory.edu} \\}

\date{}

\begin{document}
\maketitle

\begin{abstract}
As conversational AI-based dialogue management has increasingly become a trending topic, the need for a standardized and reliable evaluation procedure grows even more pressing.
The current state of affairs suggests various evaluation protocols to assess chat-oriented dialogue management systems, rendering it difficult to conduct fair comparative studies across different approaches and gain an insightful understanding of their values.
To foster this research, a more robust evaluation protocol must be set in place.
This paper presents a comprehensive synthesis of both automated and human evaluation methods on dialogue systems, identifying their shortcomings while accumulating evidence towards the most effective evaluation dimensions.
A total of 20 papers from the last two years are surveyed to analyze three types of evaluation protocols: \textit{automated}, \textit{static}, and \textit{interactive}.
Finally, the evaluation dimensions used in these papers are compared against our \textit{expert} evaluation on the system-user dialogue data collected from the Alexa Prize 2020.
\end{abstract}

\section{Introduction}

Most successful automated dialogue systems follow task-oriented dialogue management methodology, which defines an explicit goal that the system is seeking to fulfill through the conversation with the user \cite{gao2019neural}.
Recently, the research in chat-oriented dialogue management has experienced a substantial increase in popularity. Unlike task-oriented dialogues, where the success is generally measured as ability to complete the goal of the task, evaluation of chat-oriented dialogues is much less straightforward, since the conversational goals can be highly subjective \cite{huang2019challenges}. 

The evaluation of chat-oriented dialogue systems has been typically accomplished through the use of automated metrics and human evaluation (Section~\ref{sec:evaluation-protocols}). 
Automated evaluation requires no human labor once the evaluation script is written (Section~\ref{sec:analysis-automated}).
For automated evaluation to be a reliable measurement of the dialogue system quality, however, it needs to be shown to be a close approximation of human judgements (Section~\ref{sec:analysis-human}). 
Unfortunately, commonly used automated metrics correlate weakly with human judgments, indicating poor utility of such metrics \cite{liu_how_2016}. Human evaluation has become more commonplace in recent dialogue system works; however, it presents its own challenges. For one, it is time-consuming and expensive to obtain human judgments.
More critically, there is a lack of standardized protocol for such human evaluation, which makes it challenging to compare different approaches to one another. 



There have been many previous attempts at standardizing dialogue system evaluations. A major limitation has been their focus on task-oriented dialogue systems, which does not translate well to chat-oriented dialogue systems \cite{paradise,multimodal_usability}. Previous works which have included chat-oriented evaluations have lacked comprehensive coverage over the many varieties of such evaluation procedures that are currently in use. Instead, the emphasis has rested primarily on automated metrics at the expense of detailed analysis of human evaluation \cite{deriu_dialogue_survey}. At this stage in conversational AI, it is probable that automated and human metrics reveal different aspects of dialogue systems \cite{hashimoto_unify}. It would be remiss to focus on a single evaluation category when assessing the state of the field. For this reason, our work aims to fill in the gaps of previous dialogue system evaluation surveys by identifying and comparing human evaluation protocols for chat-oriented dialogue systems. 

To this end, we present a comparative analysis of the evaluations used for chat-oriented dialogue systems over the past several years. Since the field of conversational AI has experienced a rapid growth in these years, it presents a unique opportunity to observe and assess which evaluation metrics have been most widely adopted by the larger community in this period of expeditious development. We provide a detailed survey of both automated and human evaluations in order to present the most accurate depiction of the current evaluation protocols. However, our in-depth analysis is limited to that of the human evaluations due to the abundance of previous work in automated metric analysis. As such, we defer to such work as \newcite{liu_how_2016}, \newcite{ghandeharioun_approximating_2019}, and \newcite{ghazarian_better_2019} for more detail on automated metrics.

As a part of our analysis, we also present a case study of real human-machine dialogues which explores the significance of different human evaluation metrics in terms of overall user satisfaction through an expert analysis.
As a result of our work, the most commonly used evaluation metrics in contemporary literature - both automated and human - are revealed in detail and our findings towards the prevalence, impact, and applicability of human evaluation metrics are illustrated.

\section{Evaluation Protocols}
\label{sec:evaluation-protocols}

For a holistic understanding of current evaluation protocols on dialogue systems, we have carefully selected 20 relevant papers since 2018, primarily from top-tier venues, and synthesized their methods.
These papers focus on open domain (or non-task-oriented) dialogue, and employ a variety of approaches including:\footnote{Throughout the paper, the following are used to refer to the related work: 
1: \newcite{li_syntactically_2018}
2: \newcite{liu_knowledge_2018}
3: \newcite{luo_auto-encoder_2018}
4: \newcite{moghe_towards_2018}
5: \newcite{parthasarathi_extending_2018}
6: \newcite{xu_better_2018}
7: \newcite{young_augmenting_2018} 
8: \newcite{zhang_personalizing_2018}
9: \newcite{du_boosting_2019}
10: \newcite{li_incremental_2019}
11: \newcite{lin_moel_2019}
12: \newcite{madotto_personalizing_2019}
13: \newcite{qiu_are_2019}
14: \newcite{tian_learning_2019}
15: \newcite{wu_proactive_2019}
16: \newcite{zhang_recosa:_2019}
17: \newcite{zhou_unsupervised_2019}
18: \newcite{zhu_retrieval-enhanced_2019}
19: \newcite{adiwardana_towards_2020} 
20: \newcite{wang_improving_2020}. }

\begin{itemize}
    \item Incorporation of knowledge bases\newline [2, 4, 7, 18, 20]
    \item Integration of personality [8, 12]
    \item Handling of emotion-driven responses [10]
    \item Purely depending on neural-based sequence-to-sequence models [19]
\end{itemize}

\noindent Based on these papers, three main categories are found as evaluation protocols for open-domain dialogue systems: \textit{automated}, \textit{static}, and \textit{interactive}.
Automated evaluation is performed systematically by a batch script such that no human effort is required once the script is written (Section~\ref{ssec:automated-evaluation}).
Static evaluation is done by human where the evaluator assesses a dialogue whose last utterance is generated by the dialogue system (Section~\ref{ssec:static-evaluation}).
Interactive evaluation is also done by human, although the evaluator assesses the quality of the dialogue after directly interacting with the dialogue system (Section~\ref{ssec:interactive-evaluation}).

Table \ref{tab:eval} shows the distributions of the three evaluation protocols.
Most recent approaches adopt both automated and human evaluations, with only 2 papers not including any form of human evaluation.
The most common protocol for human evaluation is static evaluation, with very few papers conducting interactive assessments of dialogue systems.
No work has adopted all three types of evaluation protocols.

\begin{table}[htbp!]
\centering\small 
\begin{tabular}{c||l|c}
\bf Method & \MC{c|}{References} & \bf \# \\
\hline\hline
\multirow{2}{*}{\bf AUT} & [1, 2, 3, 4, 5, 6, 7, 9, 10, 11, 12 & \multirow{2}{*}{17} \\
                         & $\,\,$13, 14, 15, 16, 17, 20] & \\
\hline
\multirow{2}{*}{\bf STA} & [1, 3, 4, 7, 9, 10, 11, 12, 13, 14 & \multirow{2}{*}{16} \\
                         & $\,\,$15, 16, 17, 18, 19, 20] & \\
\hline
\bf INT & [8, 19] &  2 \\
\hline
\multirow{2}{*}{\bf AUT \& STA} & [1, 3, 4, 7, 9, 10, 11, 12, 13, 14 & \multirow{2}{*}{14} \\
            & $\,\,$15, 16, 17, 20] & \\
\hline
\bf AUT \& INT & [ ] & 0 \\
\hline
\bf STA \& INT & [19] & 1 \\
\end{tabular}
\caption{Distributions of the three evaluation protocols. \#: number of papers using the corresponding protocol, AUT/STA/INT: automated/static/interactive evaluation. \&: approaches using both protocols.}
\label{tab:eval}
\end{table}


\vspace{-3ex}
\subsection{Automated Evaluation}
\label{ssec:automated-evaluation}

Automated evaluation provides an objective quantitative measurement of the dialogue systems by operationalizing various dimensions of dialogue into mathematical formulations.
Depending on the specific objectives behind different systems, a few studies define novel automated metrics to capture the benefit of their proposed approaches.
Automated evaluation provides the most straightforward\LN and undemanding methods by which to evaluate dialogue systems; however, they are generally viewed as poor indicators of true dialogue quality, following results from \newcite{liu_how_2016}.


\vspace{-0.5ex}
\subsection{Static Evaluation}
\label{ssec:static-evaluation}

Static evaluation is an offline procedure where the evaluators never directly interact with the dialogue systems under review; instead, they are provided with dialogue excerpts.
These excerpts are generated by first randomly sampling dialogues from a corpus consisting of human-to-human conversations, then having the systems produce responses to the sampled dialogues.
The sampled dialogues together with the system responses are provided to human evaluators to assess.
Because only the last utterance in these excerpts are generated by the dialogue systems, it is difficult to evaluate sequential aspects about dialogue management  through static evaluation (e.g., coherence among responses generated by the same system).

\begin{table*}[tb]
\centering\resizebox{\textwidth}{!}{
\begin{tabular}{c||c|c|c|c|c|c|c|c|c|c|c|c|c|c|c|c|c|c|c|c||c}
& 1 & 2 & 3 & 4 & 5 & 6 & 7 & 8 & 9 & 10 & 11 & 12 & 13 & 14 & 15 & 16 & 17 & 18 & 19 & 20 & \# \\
\hline\hline
\texttt{BLEU} & & & \cmark & \cmark & \cmark & \cmark & & & \cmark & \cmark & \cmark & \cmark & \cmark & \cmark & \cmark & \cmark & \cmark & & & \cmark & 14 \\ \hline
\texttt{C} & & & & & & & & & & & & \cmark & & & & & & & & & 1 \\ \hline
\texttt{Coherence} & & & & & & \cmark & & & & & & & & & & & & & & & 1 \\ \hline
\texttt{Distinct} & \cmark & & \cmark & & & \cmark & & & \cmark & & & & \cmark & \cmark & \cmark & \cmark & \cmark & & & & 9 \\ \hline
\texttt{Embedding} & \cmark & & & & & & & & \cmark & & & & \cmark & \cmark & & & \cmark & & & & 5 \\ \hline
\texttt{Entity A/R} & & \cmark & & & & & & & & & & & & & & & & & & & 1 \\ \hline
\texttt{Entity Score} & & & & & & & \cmark & & & & & & & & & & & & & \cmark & 2 \\ \hline
\texttt{Entropy} & & & & & & & & & & & & & & \cmark & & & & & & & 1 \\ \hline
\texttt{Inertia} & & & & & & & & & \cmark & & & & & & & & & & & & 1 \\ \hline
\texttt{Perplexity} & & & & & \cmark & & \cmark & & & \cmark & & \cmark & & & \cmark & \cmark & & & & \cmark & 7 \\ \hline
\texttt{ROUGE} & & & & \cmark & & & & & \cmark & & & & & & & & & & & & 2
\end{tabular}}
\caption{Metrics of the automated evaluation used by recent papers on open-domain dialogue systems. The top row shows the reference numbers to the 20 surveyed papers. \#: number of papers using the corresponding metrics.}
\label{tbl:automated-evaluation}
\end{table*}


\vspace{-0.5ex}
\subsection{Interactive Evaluation}
\label{ssec:interactive-evaluation}

Unlike static evaluation, interactive evaluation has the same person play the role of both the user (one who interacts with the system) and the evaluator. 
In this setup, the evaluator has a conversation with the dialogue system and makes the assessment at the end of the conversation.
Even though this procedure is more demanding in terms of time and human effort than static evaluation, it allows the evaluator to gain a better sense of the capability of the dialogue system through explicit interaction.

\section{Analysis of Automated Evaluation}
\label{sec:analysis-automated}

Table~\ref{tbl:automated-evaluation} shows the 11 metrics used for automated evaluation in our survey:

\begin{itemize}
\setlength\itemsep{0.3em}
\item \texttt{BLEU}: a subset of BLEU-1 through BLEU-4 \cite{papineni_bleu_2002}
\item \texttt{C}: sum of entailment scores between response and persona description \cite{madotto_personalizing_2019}
\item \texttt{Coherence}: average word embedding similarity between dialogue context and generated response \cite{xu_better_2018}
\item \texttt{Distinct}: a subset of Distinct-1, Distinct-2, and Distinct-sentence \cite{li2016diversity} 
\item \texttt{Embedding}: a subset of average, extrema, and greedy embedding similarity \cite{liu_how_2016}
\item \texttt{Entity A/R}: Accuracy and recall for including the correct entities in the response \cite{liu_knowledge_2018}
\item \texttt{Entity Score}: average number of entities per response \cite{young_augmenting_2018}
\item \texttt{Entropy}: average character-level entropy over all responses \cite{mou2016sequence}
\item \texttt{Inertia}: inertia on the clusters of embeddings of responses \cite{du_boosting_2019}
\item \texttt{Perplexity}: inverse likelihood of predicting the responses of the test set \cite{chen_1998_perplexity}
\item \texttt{ROUGE}: a subset of ROUGE-1, ROUGE-2, and ROUGE-L \cite{lin_rouge_2004} 
\end{itemize}

\begin{table*}[tb]
\centering\small 
  \begin{tabular}{c||c|c|c|c|c|c|c|c|c|c|c|c|c|c|c|c||c}
    & 1 & 2 & 3 & 4 & 7 & 8 & 10 & 11 & 12 & 13 & 14 & 15 & 17 & 18 & 19 & 20 & \# \\
    \hline\hline
\tt Appropriateness     &&&&& \cmark &&&&&&&&& \cmark &&& 2 \\
\hline
\tt Coherence           &&& \cmark &&&&&&&&& \cmark &&&&& 2 \\
\hline
\tt Consistency         & \cmark &&&&& \cmark &&& \cmark &&&&&&&& 3 \\
\hline
\tt Context Coherence   &&&&&&& \cmark &&&&&&&&&& 1 \\
\hline
\tt Correctness         && \cmark &&&&&&&&&&&&&& \cmark & 2 \\
\hline
\tt Diversity           &&&&&&&&&& \cmark &&&&&&& 1 \\
\hline
\tt Emotion             & \cmark &&&&&&&&&&&&&&&& 1 \\
\hline
\tt Empathy             &&&&&&&& \cmark &&&&&&&&& 1 \\
\hline
\tt Engagingness        &&&&&& \cmark &&&&&&&&&&& 1 \\
\hline
\tt Fluency             && \cmark & \cmark & \cmark && \cmark & \cmark & \cmark & \cmark &&& \cmark &&&& \cmark & 9 \\
\hline
\tt Grammaticality      &&&&&&&&&&&&&& \cmark &&& 1 \\
\hline
\tt Humanness           &&&& \cmark &&&&&&&&&&&&& 1 \\
\hline
\tt Informativeness     &&&&& \cmark &&&&&& \cmark & \cmark && \cmark &&& 4 \\
\hline
\tt Knowledge Rel. && \cmark &&&&& \cmark &&&&&&&&& \cmark & 3 \\
\hline
\tt Logic               & \cmark &&&&&&&&&&&&&&&& 1 \\
\hline
\tt Proactivity         &&&&&&&&&&&& \cmark &&&&& 1 \\
\hline
\tt Quality             &&&&&&&&&&& \cmark && \cmark &&&& 2 \\
\hline
\tt Readability         &&&&&&&&&& \cmark &&&&&&& 1 \\
\hline
\tt Relevance           &&&& \cmark &&&& \cmark && \cmark &&&&&&& 3 \\
\hline
\tt Sensibleness        &&&&&&&&&&&&&&&  \cmark&& 1 \\
\hline
\tt Specificity         &&&& \cmark &&&&&&&&&&&  \cmark&& 2
  \end{tabular}

  \caption{Dimensions of the human evaluation used by recent dialogue system papers. The top row shows the reference numbers to the 20 survey papers. [5, 6] do not perform any human evaluation; [9, 16] perform human evaluation without reference to dimensions. \#: number of papers adopting the corresponding dimensions.}
  \label{tbl:human-evaluation}
\end{table*}

\noindent The automated metrics in Table~\ref{tbl:automated-evaluation} fall into the following five categories:

\paragraph{Ground Truth Response Similarity}
Most commonly used automated metrics focus on assessing how well system responses match the ground truth human responses, using word overlap (\texttt{BLEU}, \texttt{ROUGE}) or embedding similarity.

\paragraph{Context Coherence}
Embedding similarities between dialogue contexts and system responses have been used to quantitatively assess the relevance between the system responses and the preceding dialogue history (\texttt{Coherence}, \texttt{Embedding}).

\paragraph{Response Diversity}
Other widespread metrics assess the diversity of the system responses in order to determine the amount of repetition and generic content in the system responses (\texttt{Distinct}, \texttt{Entropy}, \texttt{Inertia}, \texttt{Entity Score}).

\paragraph{Language Model Fitness}
Generative models are usually evaluated in terms of how well they learn to model the language of the dialogues in their training corpus (\texttt{Perplexity}).

\paragraph{Application-Specific}
The other observed metrics can be considered application-specific since \texttt{Entity A/R} is used to measure the ability of the system to produce the correct entities in its responses and \texttt{C} is specifically created as a measure of the consistency between the dialogue responses and their respective persona descriptions.



\section{Analysis of Human Evaluation}
\label{sec:analysis-human}

While automated evaluation measures dimensions of dialogue objectively, human evaluation captures the subjective assessment from the user's point of view. 
Regardless of the exact method chosen, all human evaluations involve gathering external annotators who answer questions regarding the dialogues resulting from a dialogue system.


\subsection{Dimensions of Human Evaluation}

There is high variability in the dimensions of dialogue that previous studies have used for assessing dialogue systems in both static and interactive evaluations.
Table~\ref{tbl:human-evaluation} provides a detailed overview of the dimensions used by each of the surveyed papers when evaluating their work.
There are a total of 21 uniquely-worded dimensions found; 11 of them appear in only a single paper.
The resulting matrix provides clear evidence of the inconsistencies in human evaluation methods, as its sparsity is indicative of low overlap among those methods.
The long tail distribution of the evaluation metrics makes it difficult for cross-work comparisons without a substantial study to align the disparate evaluation of one work with another. 

\noindent Although the evaluation dimensions appear to be distinct on the surface, several of them appear to be similar in meaning. 
To analyze the level of overlap among the seemingly distinct evaluation dimensions, we compile the definitions and instructions shared by each of the papers regarding their evaluation dimensions and rating scales. 
Based on manual analysis, we are able to group dimensions together that are indeed evaluating the same aspect of dialogue as one another, even though the authors mention them by different names.
Table~\ref{tbl:dimension-groups} provides the dimension groupings that are identified on the basis of their respective definitions.

\begin{table*}[htbp!]
\centering

\begin{subtable}{\textwidth}
\centering\small
    \begin{tabular}{c|l}
    \hline
    \multirow{4}{*}{\tt Fluency}
    & Whether the response from the listener is understandable \cite{lin_moel_2019} \\ \cline{2-2}
    & Whether the response is fluent and natural \cite{li_incremental_2019} \\ \cline{2-2}
    & Whether each sentence has correct grammar \cite{luo_auto-encoder_2018} \\ \cline{2-2}
    & Fluency measures if the produced response itself is fluent \cite{wu_proactive_2019}: \\
    \hline
    \tt Consistency
    & Whether the reply is fluent and grammatical \cite{li_syntactically_2018} \\ \hline
    \tt Readability
    & Whether the utterance is grammatically formed \cite{qiu_are_2019} \\ \hline
    \tt Grammaticality
    & Whether the response is fluent and grammatical \cite{zhu_retrieval-enhanced_2019} \\ \hline
    \end{tabular}
\caption{Grammatical Capability.}
\label{tab:grammar}
\vspace{0.5ex}
\end{subtable}

\begin{subtable}{\textwidth}
\centering\resizebox{\textwidth}{!}{
    \begin{tabular}{c|l}
    \hline
    \multirow{3}{*}{\tt Relevance}
        & Whether the responses of the listener seem appropriate to the conversation \cite{lin_moel_2019} \\ \cline{2-2}
        & Whether the response is appropriate/relevant in the current context language \cite{moghe_towards_2018} \\ \cline{2-2}
        & Whether the reply is relevant to the query \cite{qiu_are_2019} \\ \hline
    \tt Appropriateness
        & Whether the response is appropriate in grammar, topic, and logic \cite{young_augmenting_2018} \\ \hline
    \multirow{2}{*}{\tt Coherence}
        & Whether the generated response is relevant to the input \cite{luo_auto-encoder_2018} \\ \cline{2-2}
        & Whether the whole dialogue is fluent (does not contain irrelevant or illogical responses)  \cite{wu_proactive_2019} \\
         \hline
    \tt Context Coherence
        & Whether the response is coherent with the context and guides the following utterances \cite{li_incremental_2019} \\ \hline
    \tt Logic
        & Whether the post and the reply are logically matched \cite{li_syntactically_2018} \\\hline
    \tt Sensibleness
        & Whether the response makes sense given the context \cite{adiwardana_towards_2020} \\ \hline
    \end{tabular}}
\caption{Turn Coherence.}
\label{tab:coherence}
\vspace{0.5ex}
\end{subtable}

\begin{subtable}{\textwidth}
\centering\resizebox{\textwidth}{!}{
    \begin{tabular}{c|l}
    \hline
    \multirow{4}{*}{\tt Informativeness}
        & Whether the response provides new information and knowledge in addition to the post \cite{young_augmenting_2018} \\ \cline{2-2}
        & Whether the response has unique words and multi-topic clauses \cite{tian_learning_2019} \\ \cline{2-2}
        & Whether the response has meaningful information relevant to its message \cite{zhu_retrieval-enhanced_2019} \\ \cline{2-2}
        & Whether the model makes full use of knowledge in the response \cite{wu_proactive_2019}\\ \hline
    \multirow{2}{*}{\tt Specificity}
        & Whether the model produced movie-specific responses or generic responses \cite{moghe_towards_2018} \\ \cline{2-2}
        & Whether the response is specific to the context \cite{adiwardana_towards_2020} \\ \hline
    \tt Diversity
        & Whether the reply narrates with diverse words \cite{qiu_are_2019} \\ \hline
    \end{tabular}}
\caption{Response Informativeness.}
\label{tab:specificity}
\end{subtable}

\caption{Proposed reductions of dialogue evaluation dimensions into non-overlapping components}
\label{tbl:dimension-groups}
\end{table*}

Definitions in Table \ref{tab:grammar} aim to address the grammaticality of system responses, including words like \textit{grammar}, \textit{understandable}, and \textit{accurate}. 
As a result, the four dimensions recorded in this table can be viewed as lexical variations of the same underlying \texttt{Grammaticality} dimension.
Similarly, definitions in Table \ref{tab:coherence} highlight keywords like \textit{appropriate}, \textit{relevant}, and \textit{on-topic}, thus providing evidence that each of those dimensions are instances of the \texttt{Relevance} dimension. 
Finally, Table \ref{tab:specificity} has a high occurrence of information and diversity-focused definitions, and we can reduce the dimensions shown there to the single \texttt{Informativeness} dimension. 

Other than these highly overlapping dimensions, \texttt{Quality} \cite{tian_learning_2019,zhou_unsupervised_2019} and \texttt{Humanness} \cite{moghe_towards_2018} can both be considered as the single \texttt{Quality} dimension, since they are used to elicit an overall quality assessment of the dialogue system responses. 
Similarly, \texttt{Emotion} \cite{li_syntactically_2018} and \texttt{Empathy} \cite{lin_moel_2019} can be reduced into the \texttt{Emotional Understanding} dimension that captures both the comprehension and production of emotional responses. 
The remaining two dialogue dimensions assess a unique quality of dialogue and are useful as independent dialogue dimensions:

\begin{itemize}
	\item \texttt{Engagingness}: whether the response includes interesting content \cite{zhang_personalizing_2018}
	\item \texttt{Proactivity}: whether the response introduces new topics without breaking coherence \cite{wu_proactive_2019}
\end{itemize}

\noindent Finally, two evaluation dimensions are specifically used for a subset of dialogue systems that incorporate knowledge: 

\begin{itemize}
	\item \texttt{Correctness}: was the response accurate based on the real-world knowledge \cite{liu_knowledge_2018,wang_improving_2020}
	\item \texttt{Knowledge Relevance}: was the knowledge shared in the response appropriate to the context \cite{liu_knowledge_2018,wang_improving_2020}
\end{itemize}

\noindent \texttt{Knowledge Relevance} is very similar to the previously discussed \texttt{Relevance} dimension, although it is specifically targeting an assessment of the appropriateness of the knowledge being used. 
Even more niche, the \texttt{Correctness} dimension is unique to knowledge-focused systems that seek to present only true factual information to the user; thus, such a dimension may not be useful in other contexts. 
Due to their targeted nature, these two dimensions may fall outside of the scope of a general, comprehensive, unified evaluation of dialogue systems, and instead be used for a targeted subgroup.

\begin{table*}[htbp!]
\centering\resizebox{\textwidth}{!}{
  \begin{tabular}{l||l}
  \bf Dimension & \bf Definition\\
  \hline\hline
  \tt Grammaticality & Responses are free of grammatical and semantic errors \\
  \tt Relevance & Responses are on-topic with the immediate dialogue history \\
  \tt Informativeness & Responses produce unique and non-generic information that is specific to the dialogue context \\
  \tt Emotional & Responses indicate an understanding of the user's current emotional state and \\
  \tt Understanding & provide an appropriate emotional reaction based on the current dialogue context  \\
  \tt Engagingness & Responses are engaging to user and fulfill the particular conversational goals implied by the user \\
  \tt Consistency & Responses do not produce information that contradicts other information known about the system \\
  \tt Proactivity & Responses actively and appropriately move the conversation along different topics \\
  \tt Quality & The overall quality of and satisfaction with the dialogue \\
  \end{tabular}}
  \caption{The final set of our proposed dialogue dimensions for human evaluation.}
  \label{tab:dims}
\end{table*}

\noindent In total, after merging similar dimensions and discarding non-generalizable dimensions, a total of eight dimensions have been identified that share little to no definitional overlap and are reasonably applicable to all dialogue systems. 
Table \ref{tab:dims} shows the finalized set of dialogue evaluation dimensions.

\subsection{Diversities in Evaluation Metrics}

Aside from the discrepancies in dialogue dimensions used for evaluation among different works, the actual procedure of evaluating these dialogue dimensions varies even further, particularly for static evaluations.
A majority of work instructs human annotators to rate the dialogue system responses on a set of dialogue dimensions using numeric scales, where the scales being used are often different even between works that employ the same dialogue dimensions.  
For instance, one of the most commonly used dimension is the \texttt{Fluency} of the dialogue, with 9 out of the 16 papers in Table~\ref{tbl:human-evaluation} have adopted this as an evaluation dimension.
Between those 9 studies, \texttt{Fluency} ratings include scales of:

\begin{itemize}
	\item \texttt{0$\sim$2}: \newcite{wu_proactive_2019,li_incremental_2019}
	\item \texttt{0$\sim$3}: \newcite{wang_improving_2020,liu_knowledge_2018}
	\item \texttt{1$\sim$5}: \newcite{moghe_towards_2018,zhang_personalizing_2018,lin_moel_2019,madotto_personalizing_2019}
	\item \texttt{1$\sim$10}: \newcite{luo_auto-encoder_2018}
\end{itemize}


\noindent Furthermore, some studies use a preference metric for static evaluation in addition to - or even instead of - the numerical ratings \cite{lin_moel_2019,young_augmenting_2018,du_boosting_2019,zhang_recosa:_2019}. 
In this case, human annotators are asked to select the most compelling response among many generated by multiple dialogue systems or even humans.
Thus, preference metrics provide estimated ranking scores among different systems by measuring the percentage of times each system is preferred over the others.

\noindent Unlike the diversity in static evaluation, for the two papers, \newcite{zhang_personalizing_2018} and \newcite{adiwardana_towards_2020}, employing interactive evaluation, only numerical ratings on specific dialogue dimensions are used as evaluation methods; other methods such as preference metrics are not used in either case.


\subsection{Static vs Interactive Evaluations}

Establishing the necessary assessment metrics is only one consideration to achieve an accurate dialogue evaluation. 
The other major consideration is the procedure underlying the evaluation. 
This section discusses the two human evaluation protocols, static and interactive evaluations, that have previously been used by many dialogue systems.

\noindent Although both evaluation protocols overcome the deficiencies brought forth by automated evaluation through human judgment, interactive evaluation is hypothesized to be a more reliable assessment strategy than static one. 
What static evaluation offers above interactive evaluation is a lower cost in terms of time and labor. 
By removing the human annotator from the task of interacting with the dialogue system, and instead having them review a dialogue excerpt, the amount of work required is reduced. 

However, this is simultaneously a point in favor of static evaluation, but also a factor as to why it is less reliable. 
As \newcite{ghandeharioun_approximating_2019} suggest, chat-oriented dialogues have a less defined conversational goal which can best be summarized as being able to hold a ``natural social interaction with humans''.
The success - or failure - at this can only be evaluated by the targeted recipient of the conversation; namely, the user that the system is interacting with. External annotators, at best, can estimate the user's satisfaction with the conversation based on their own projected opinions, which is not necessarily the most accurate assessment. 

\begin{table*}[htb!]
\centering
\begin{subtable}{\textwidth}
    \centering\small
    \begin{tabular}{c||c|c|c|c|c|c|c}
    \bf OQ & \bf GR & \bf RE & \bf IN & \bf EU &  \bf EN  & \bf CO & \bf PR \\
    \hline\hline
    1 & 5.00 ($\pm$0.00) & 1.94 ($\pm$0.98) & 2.86 ($\pm$1.29) & 1.00 ($\pm$0.00) & 2.33 ($\pm$0.89) & 4.94 ($\pm$0.23) & 1.64 ($\pm$0.87)   \\
    2 & 4.70 ($\pm$0.47) & 2.85 ($\pm$0.88) & 3.25 ($\pm$1.25) & 1.15 ($\pm$0.37) & 3.15 ($\pm$0.75) & 4.90 ($\pm$0.31) & 2.15 ($\pm$0.59)   \\
    3 & 4.62 ($\pm$0.51) & 3.46 ($\pm$0.52) & 2.92 ($\pm$0.86) & 1.08 ($\pm$0.28) & 2.92 ($\pm$0.49) & 4.77 ($\pm$0.44) & 2.38 ($\pm$0.65)   \\
    4 & 4.71 ($\pm$0.46) & 3.89 ($\pm$0.42) & 4.25 ($\pm$0.70) & 1.11 ($\pm$0.31) & 3.86 ($\pm$0.36) & 4.82 ($\pm$0.39) & 2.93 ($\pm$0.54)   \\
    5 & 4.33 ($\pm$0.58) & 4.33 ($\pm$0.58) & 3.67 ($\pm$0.58) & 1.33 ($\pm$0.58) & 4.00 ($\pm$0.00) & 5.00 ($\pm$0.00) & 3.00 ($\pm$0.00) 
    \end{tabular}
    \caption{The \texttt{OQ} column shows the overall quality ratings from our expert and the other columns show the average ratings from the expert on the corresponding dialogue dimensions.}
    \label{tab:static_avgs}
    \vspace{0.5ex}
\end{subtable}

\begin{subtable}{\textwidth}
    \centering\small
    \begin{tabular}{c||c|c|c|c|c|c|c}
    \bf OQ & \bf GR & \bf RE & \bf IN & \bf EU &  \bf EN  & \bf CO & \bf PR \\
    \hline\hline
    1 & 4.85 ($\pm$0.37) & 2.20  ($\pm$1.20) & 2.95  ($\pm$1.28) & 1.00 ($\pm$0.00) & 2.60 ($\pm$1.05) & 4.85 ($\pm$0.37) & 1.95 ($\pm$0.94)   \\
    2 & 4.80 ($\pm$0.41) & 3.05  ($\pm$1.10) & 3.95  ($\pm$1.19) & 1.25 ($\pm$0.44) & 3.30 ($\pm$0.92) & 5.00 ($\pm$0.00) & 2.10 ($\pm$0.79)   \\
    3 & 4.85 ($\pm$0.37) & 2.75  ($\pm$1.07) & 2.50  ($\pm$0.95) & 1.00 ($\pm$0.00) & 2.60 ($\pm$0.75) & 4.90 ($\pm$0.31) & 2.05 ($\pm$0.89)   \\
    4 & 4.65 ($\pm$0.49) & 3.40  ($\pm$0.82) & 3.30  ($\pm$0.92) & 1.10 ($\pm$0.31) & 3.25 ($\pm$0.79) & 4.85 ($\pm$0.37) & 2.25 ($\pm$0.72)   \\
    5 & 4.80 ($\pm$0.41) & 3.30  ($\pm$1.13) & 4.10  ($\pm$0.97) & 1.05 ($\pm$0.22) & 3.50 ($\pm$0.76) & 4.80 ($\pm$0.41) & 2.85 ($\pm$0.75) 
    \end{tabular}
    \caption{The \texttt{OQ} column shows the overall quality ratings from the Alexa Prize and the other columns show the average ratings from the expert on the corresponding dialogue dimensions.}
    \label{tab:interactive_avgs}
    \vspace{0.5ex}
\end{subtable}

\caption{The average ratings by our expert on each of the dialogue dimensions in Table~\ref{tab:dims} with respect to the overall ratings from the expert and the Alexa Prize. \texttt{OQ}: Quality, \texttt{GR}: Grammaticality, \texttt{RE}: Relevance, \texttt{IN}: Informativeness, \texttt{EU}: Emotional Understanding, \texttt{EN}: Engagingness, \texttt{CO}: Consistency, \texttt{PR}: Proactivity.}
\label{tbl:expert-ratings}
\end{table*}

In addition, static evaluation is commonly conducted by producing a single system response in a fixed dialogue context. 
This fails to reveal certain system deficiencies, such as repetitiveness, inconsistency, and lack of long-term memory of the information shared in the conversation. 
It also prevents an assessment of the system's error-handling or misunderstanding recovery capabilities from being encountered. All of these aspects are necessary to truly assess the quality of dialogues that a given dialogue system can produce. Without this information, only a biased perspective can be achieved, and the evaluation will not reflect the true capability of the system if it were to be used in practice.

\section{Case Study: Alexa Prize 2020}
\label{sec:alexa-prize}

This section presents a case study of the significance of the proposed dialogue dimensions in Table~\ref{tab:dims} using real human-machine dialogues.
For this analysis, 100 rated conversations were taken from the Alexa Prize Socialbot Grand Challenge 3\footnote{\url{https://developer.amazon.com/alexaprize}}, which is a university competition to create innovative open-domain chatbots \cite{alexaprize}. During the competition, conversations are rated in terms of \texttt{Overall Quality} on a scale of \texttt{1} \texttt{(worst)}  to \texttt{5} \texttt{(best)} under the interactive evaluation protocol.
For this case study, we sampled conversations with an equal distribution between all ratings, where every conversation has at least three turns to ensure sufficient content. 

Because only the \texttt{Overall Quality} dimension is provided from the interactive evaluation, we also conducted an expert analysis on the same conversations in order to explore the implications of the other previously identified dialogue dimensions. To this end, one of the authors - who has over three years of experience in dialogue system research - manually rated the conversations on each of the dialogue dimensions in Table~\ref{tab:dims}.

It is worth mentioning that the following findings are taken as only a preliminary analysis, strongly considering the low agreement between the expert and interactive evaluations on \texttt{OQ}, which will be discussed shortly (Section~\ref{ssec:expert-vs-interactive}). 
This disparity between the expert and human user evaluations renders it difficult to convey a convincing conclusion regarding the significance of the evaluation dimensions.
However, we hope this work begins the momentum to investigate the importance of such evaluation dimensions in overall human perception of dialogue quality.


\subsection{Quality vs. Other Dialogue Dimensions}

Table~\ref{tbl:expert-ratings} shows the average rating and its standard deviation on each of the 7 dialogue dimensions (\texttt{GR}, \texttt{RE}, \texttt{IN}, \texttt{EU}, \texttt{EN}, \texttt{CO}, \texttt{PR}) across the overall quality ratings (\texttt{OQ}).
All ratings on those 7 dimensions are assessed by our expert.
\texttt{OQ} ratings are provided by the expert for Tables~\ref{tab:static_avgs} and the human users from the Alexa Prize for Table~\ref{tab:interactive_avgs}.


\paragraph{Relevance \& Proactivity} 
The clearest positive relationship to \texttt{OQ} is observed from \texttt{RE} and \texttt{PR}, especially from the expert evaluation although it can be seen in the interactive evaluation as well. This suggests that these dimensions are pertinent to the human perception of dialogue quality, and that this relationship is even more apparent when evaluators are given the opportunity to review previous dialogue turns when determining \texttt{OQ}.

\paragraph{Informativeness \& Engagingness} 
The relation ship between \texttt{IN} and \texttt{EN} to \texttt{OQ} is not as obvious as the previous two dimensions, \texttt{RE} and \texttt{PR}, although an indication of a positive relationship is observed.

\paragraph{Grammaticality} 
Due to the manual curation of responses in our Alexa Prize chatbot, we have tight control over the grammaticality of our responses; thus, the overall variance in \texttt{GR} is low. 
Interestingly, we do notice that there is a slight inverse relationship between \texttt{GR} and \texttt{OQ}. 
Although this may seem counter-intuitive, the likely explanation is that conversations with higher \texttt{OQ} tend to be longer so that they comprise a greater number of topics and, as more topics are introduced, the chance for an (accidentally) ungrammatical response to be revealed is higher. 
Nonetheless, it appears that ungrammaticality is not a strict deterrent on \texttt{OQ}.

\paragraph{Emotional Understanding \& Consistency} 
The effect of \texttt{EU} and \texttt{CO} on \texttt{OQ} is inconclusive from the presented analysis.
This is attributed to the low variation in these dimensions of our chatbot, as we can enforce the consistency of responses and do not aim to tackle emotional understanding.


\subsection{Expert vs.\ Interactive Evaluations}
\label{ssec:expert-vs-interactive}

The inter-annotator agreement between the \texttt{OQ} ratings of the expert and the users from the Alexa Prize is provided in Table~\ref{tab:cohen}.
The agreement is measured for both fine-grained ratings that consider all scales (\texttt{1} - \texttt{5}) and coarse-grained ratings that consider only two scales (low: \texttt{1} - \texttt{2}, high: \texttt{3} - \texttt{5}).
Although the inter-annotator agreement is higher for the coarse-grained ratings, it is apparent that the agreement scores are dramatically low for both.

\begin{table}[htbp!]
\centering\small 
\begin{tabular}{c|c}
\bf Rating Type & \bf Agreement \\
\hline\hline
\bf Fine-grained   & 0.13 \\
\bf Coarse-grained & 0.22 \\
\end{tabular}
\caption{Cohen's Kappa scores on the overall quality ratings between the expert and interactive evaluation.}
\label{tab:cohen}
\end{table}

\noindent Table~\ref{tab:dist} shows that the expert evaluation tends to be more punishing overall, with a much fewer number of conversations receiving a \texttt{5.0} rating. Indeed, 56\% of the conversations from the expert evaluation would be categorized as a low rating, whereas the interactive evaluation has only 40\%. 
Even so, the low agreement indicates that the quality assessments across the two evaluation protocols are highly variable across the same conversations.

\begin{table}[htbp!]
\centering\small 
\begin{tabular}{c||c|c|c|c|c||c}
\tt OQ & \tt 1 & \tt 2 & \tt 3 & \tt 4 & \tt 5 & $\sum$ \\
\hline\hline
\bf Interactive & 20 & 20 & 20 & 20 & 20 & 100 \\
\bf Expert & 36 & 20 & 13 & 28 & 3 & 100 \\
\end{tabular}
\caption{Comparison of the rating distribution between expert and interactive evaluation}
\label{tab:dist}
\end{table}

\noindent This provides preliminary support for the hypothesis in Section~\ref{sec:analysis-human} that external evaluators are unable to accurately infer the same impression of a conversation as that of the user who is actually participating in the conversation. Although there are potential methods which aim to mitigate this effect - such as agglomerate ratings across more than one external annotator - the underlying cause of such variance may be attributed to the poor suitability of external evaluations for dialogue system evaluation as a whole, but further work is required.

\section{Conclusion and Future Work}

In this paper, we provide an extensive background and the current states on the three types of dialogue system evaluation protocols, automated, static, and interactive. 
Our analysis shows that static evaluation is the dominating human evaluation used in the most recent dialogue system works, although it has several concerning limitations, some of which are exemplified through our case study. 
We propose a set of eight dialogue dimensions that encapsulate the evaluations of previous studies without redundancy. 
As a result of our case study, we find preliminary evidence that the dimensions of relevance, proactivity, informativeness, and engagingness are likely to be contributing factors to the overall perception of dialogue quality.  

Our future work will build upon these findings to develop a thorough understanding of the necessary dialogue dimensions for comprehensive interactive evaluation of dialogue systems. 
Through an analysis based on large-scale user studies, we look to propose an evaluation protocol that captures the human judgement of dialogue quality through precise formulation of evaluation dimensions, in order to enable targeted dialogue system advancements.

\section*{Acknowledgments}

We gratefully acknowledge the support of the Alexa Prize Socialbot Grand Challenge 3.
Any contents in this material are those of the authors and do not necessarily reflect the views of the Alexa Prize.

\bibliography{acl2020}
\bibliographystyle{acl_natbib}

\end{document}